\documentclass[a4paper,fleqn]{cas-dc}
\usepackage{amssymb}
\usepackage{latexsym}
\usepackage{subfigure}
\usepackage{graphicx}
\usepackage{url}
\usepackage{xcolor}
\usepackage{color}

\usepackage{framed,multirow}
\usepackage{bm}
\usepackage{lineno,hyperref}
\usepackage{algorithm}
\usepackage{algorithmic}
\usepackage{setspace}
\usepackage[numbers]{natbib}
\def\tsc#1{\csdef{#1}{\textsc{\lowercase{#1}}\xspace}}
\tsc{WGM}
\tsc{QE}
\begin{document}
\let\WriteBookmarks\relax
\def\floatpagepagefraction{1}
\def\textpagefraction{.001}

%\shorttitle{<short title of the paper for running head>}    

%\shortauthors{<short author list for running head>}  

\title [mode = title]{Attention-based Feature Decomposition-Reconstruction Network for Scene Text Detection}  


%\tnotetext[1]{The second title footnote which is alonger text matter to fill through the whole text} 


\author[1]{Qi Zhao}
\ead{zhaoqi@buaa.edu.cn}
\author[1]{Yufei Wang}[style=chinese, orcid=0000-0003-1543-3071]
\ead{sy1902316@buaa.edu.cn}
\author[1]{Shuchang Lyu}
\ead{lyushuchang@buaa.edu.cn}
\author[1]{Lijiang Chen}
\ead{chenlijiang@buaa.edu.cn}
\cormark[1]
\affiliation[1]{organization={Department of Electrics and Information Engineering, Beihang}, 
            city={Beijing},
            postcode={100191}, 
            state={Beijing},
            country={China}}
% Footnote of the second author
% Email id of the second author

%\credit{data curation}

\cortext[1]{Corresponding author}



\begin{abstract}
Recently, scene text detection has been a challenging task. Texts with arbitrary shape or large aspect ratio are usually hard to detect. Previous segmentation-based methods can describe curve text more accurately but suffer from over segmentation and text adhesion. In this paper, we propose attention-based feature decomposition-reconstruction network for scene text detection, which utilizes contextual information and low-level feature to enhance the performance of segmentation-based text detector. In the phase of feature fusion, we introduce cross level attention module to enrich contextual information of text by adding attention mechanism on fused multi-scaled feature. In the phase of probability map generation, a feature decomposition-reconstruction module is proposed to alleviate the over segmentation problem of large aspect ratio text, which decomposes text feature according to their frequency characteristic and then reconstructs it by adding low-level feature. Experiments have been conducted on two public benchmark datasets and results show that our proposed method achieves state-of-the-art performance.
\end{abstract}


\begin{highlights}
\item We propose a decomposition-reconstruction structure, which extracts high frequency feature in the decomposition operation and fuses low-level feature in the reconstruction operation. It achieves better or competitive performance on detecting adjacent text instances.

\item We introduce a cross level attention module to obtain contextual information at different feature levels, which enables our method to handle the problem of over segmentation when the text instances have extreme aspect ratios.

\item We conduct experiments on two benchmark datasets. Experimental results validate the performance improvement of our method, especially on texts with large aspect ratios 
and curved texts. It is worth noting that we achieves the state-of-the-art performance on MSRA-TD500 dataset by a large margin.
\end{highlights}

% Keywords

\maketitle

% Main text
\section{Introduction}
\label{sec:intro}
Scene text detection has wide range real-world applications such as scene understanding, instant translation and automatic driving. Benefiting from the development of deep learning methods based on CNN, scene text detection has made great progress. Recent scene text detection methods mostly follow the problem setting of object detection or semantic segmentation. By modifying the architecture of efficient CNN models like Faster R-CNN~\cite{fasterrcnn}, SSD~\cite{ssd} and FCN~\cite{fcn}, scene text detectors achieve remarkable results on several benchmark datasets. However, due to the complexity of the text in the real-world such as extreme aspect ratio, deformed or arbitrary shape, current scene text detectors do not perform well and there is still much room for improvement. Recently, segmentation-based methods become popular due to its strong description ability of arbitrary shaped text.

\begin{figure}[!t]
\centering
\includegraphics[scale=.5]{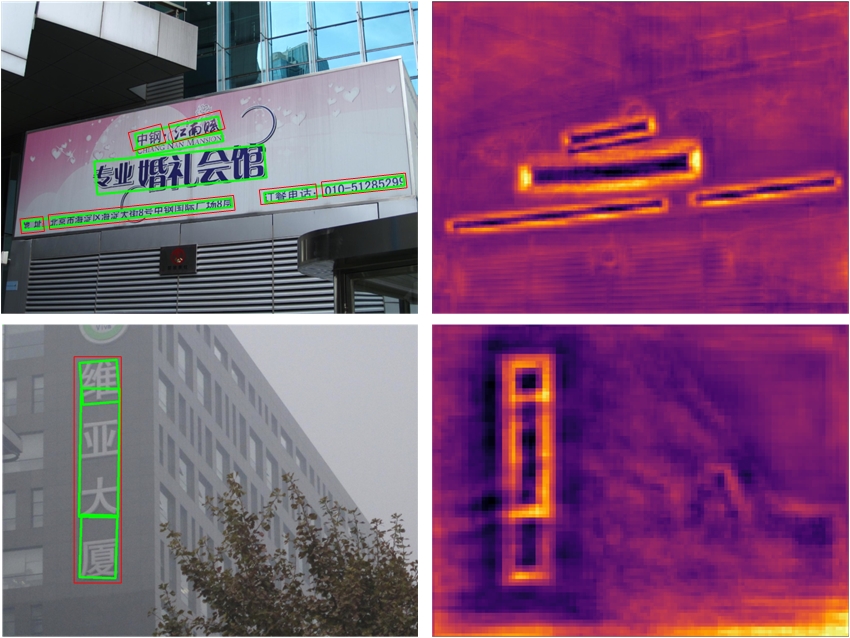}
\caption{Two common challange of segmentation-based text detectors. The first row shows the problem of text adhesion and the second row shows the problem of over segmentation. In the first column, green bounding box is the detection results and wrong cases are revised by red bounding box. The second column is visualization of intermediate feature map.}\label{fig:bad_sample}
\end{figure}

Based on a large number of prior experiments and intuitive observations, we find two common problems of segmentation-based text detectors. (1) As shown in Figure.\ref{fig:bad_sample}, it is difficult to distinguish different text instances when they are closely placed. (2) Text instances with large aspect ratios tend to have problem of over segmentation as the character spacing is too large. To address the first issue, we argue that previous methods fail to pay enough attention to the edge information between text instances, which is essential to separate adjacent texts. For the second problem, we consider that lack of contextual information causes performance degradation.

Previous notable works like DB~\cite{DB} and PAN~\cite{PAN} make effort to separate adjacent text by using threshold map or text kernels during train phase to represent the edge of a text instance or the center of text region. Such methods achieve good results but neglect the problem of over segmentation. In this work, we propose an attention-based feature decomposition-reconstruction network, which aims to jointly handle the problem of over segmentation and text adhesion. Specifically, we introduce cross level attention module and feature decomposition-reconstruction module. The cross level attention module is a series connection of channel attention and spatial attention, which is applied to the fused feature generated by FPN~\cite{FPN}. With the help of attention mechanism, features of different scales and positions are enhanced or suppressed selectively. Such operation alleviate the over segmentation problem by highlighting the high-level features, which makes detector focus more on the integrity of long text. In addition, with more contextual information involved, text detector will less likely to treat character spacing as text boundary. The feature decomposition-reconstruction module aims to solve the problem of text adhesion. This module decomposes fused feature into two kinds of features with different frequency characteristics, the high term and the low frequency term. We consider the high frequency term to represent features of text boundary and low frequency term to represent features inside the text region. The key to distinguish adjacent text is to highlight boundary region, so we add low-level feature to low frequency term to enrich boundary information. The low frequency term is generated by warping each pixel towards text inner parts through a learned offset field, and the high frequency term is obtained by explicitly subtracting the low frequency term from the input feature. Finally we merge the two kinds of features into the final representation for text region. Our method improves the accuracy of segmentation-based method DB~\cite{DB} by 2.2\% on MSRA-TD500~\cite{td500} dataset and 1.3\% on Total-Text~\cite{Totaltext} dataset.

The main contributions of this paper are as follows:
\begin{itemize}
\item We propose a decomposition-reconstruction structure, which extracts high frequency feature in the decomposition operation and fuses low-level feature in the reconstruction operation. It achieves better or competitive performance on detecting adjacent text instances.

\item We introduce a cross level attention module to obtain contextual information at different feature levels, which enables our method to handle the problem of over segmentation when the text instances have extreme aspect ratios.

\item We conduct experiments on two typical benchmark datasets. Experimental results validate the performance improvement of our method, especially on texts with large aspect ratios 
and curved texts. It is worth noting that we achieves the state-of-the-art performance on MSRA-TD500 dataset by a large margin.
\end{itemize}

\section{Related work}
\label{sec:RW}

Recently, scene text detectors based on deep learning have proved their effectiveness. Most of these methods can be divided into two categories: regression based methods and segmentation based methods.
\subsection{Regression based methods.} 
Regression based methods often take advantage of popular object detectors such as SSD~\cite{ssd} and YOLO~\cite{yolo}. These methods directly predict bounding box coordinates of text instances so that they can get rid of complex post-processing algorithm. TextBoxes~\cite{Textboxes} modifies the shape of convolutional kernels and anchor scales to capture texts with various aspect ratios. TextBoxes++~\cite{Textboxes++} uses quadrangles as text representation instead of horizontal bounding boxes to better detect arbitrary shaped text. EAST~\cite{EAST} directly regresses the coordinates of bounding box in pixel level then uses NMS to filter redundant boxes. RRD~\cite{RRD} decomposes text detection into two tasks while using rotation-invariant features for text classification and rotation-sensitive features for text location respectively. DMPNet~\cite{DMPNet} proposes using quadrilateral sliding window  as default anchor to match multi-oriented text. However, when confronted with irregular shaped text, their performance often degrade dramatically due to the limited representation capability of points sequences.
\subsection{Segmentation based methods.} 
These methods usually regard text detection as a two-stage task, which first detect fragments of text lines and then combine these fragments into final outputs. PixelLink~\cite{PixelLink} perform per-pixel prediction on text/non-text and link relation first and then uses connected components based method to obtain final detection. PSENet~\cite{PSENet} adopt FCN~\cite{fcn} structure to predict multiple scales text kernel, then uses progressive scale expansion algorithm to generate final predictions and distinguish adjacent text instances. SegLink~\cite{SegLink} and SegLink++~\cite{SegLink++} predict all segments within a text line and then learn link relation between segments to form final bounding boxes. TextSnake~\cite{Textsnake} presents a more general and flexible text region representation which consider text instance as a group of round panel along the center line of a text. SAE~\cite{SAE} learns a mapping from text pixels to embedded feature space and obtain output mask by pixel clustering. DB~\cite{DB} predicts per-pixel masks named threshold map to enhance the supervision to text boundary while increasing inference speed.  CRAFT~\cite{CRAFT} focus on the center of character region and uses Gaussian heat map to concatenate characters into text line.

\section{Methodology}
\label{sec:Method}

\begin{figure*}[!t]
\centering
\includegraphics[scale=0.7]{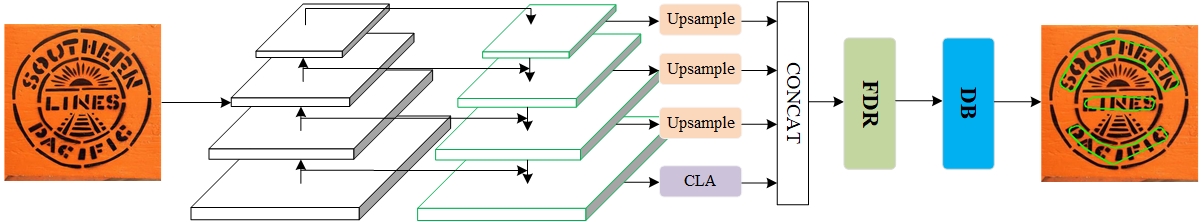}
\caption{The framework of our proposed method. Given an input image, we first use ResNet-50 with FPN to get multi-scale feature map, then we apply CLA on the feature map with largest resolution. All feature maps are then concatenated together and fed to FDR module, and finally we use DB to generate detection result.}
\label{fig:frame}
\end{figure*}

The pipeline of our proposed model is shown in Figure.\ref{fig:frame}. It is composed of a CNN backbone with a feature pyramid structure~\cite{FPN}, a cross level attention module, a feature decomposition-reconstruction module, and a differentiable binarization head.

\subsection{Network Architecture}
To validate that our proposed method can improve the performance of segmentation-based scene text detector, we adopt DB\cite{DB} structure as our segmentation head. The backbone of our model utilizes deformed ResNet-50, which applies deformable convolutions in all the $3\times3$ convolution layers in stages conv3, conv4 and conv5 of ResNet-50. The output of ResNet-50 from stage 1 to stage 4 is then fed into a feature-pyramid structure to get fused feature map with the size of $\frac{H}{4}\times\frac{W}{4}\times C$, where H and W are the height and width of the input image; $C$ is the channel number of feature map and is set to 256. Then we use cross level attention module to capture contextual information at different feature levels. After that we apply feature decomposition-reconstruction module which aims at extracting feature with high frequency and then fusing fine-detailed low-level feature to help distinguish the border region of text instances. Finally the mixed feature with rich contextual information is used to predict both the probability map and the threshold map following the pipeline introduced in DB.The structure of DB is shown in Figure.~\ref{fig:FDR-DB}

\begin{figure}[!t]
\centering
\includegraphics[scale=0.65]{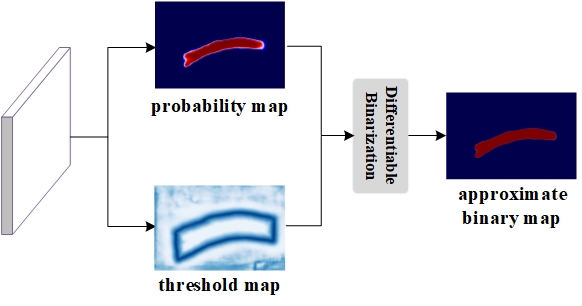}
\caption{Illustration of feature differentiable binarization. The output feature map of FDR is used to predict both probability map and threshold map. After that, we use these two maps to generate approximate binary map.}\label{fig:DB}
\end{figure}

\begin{figure*}[!t]
\centering
\includegraphics[scale=0.9]{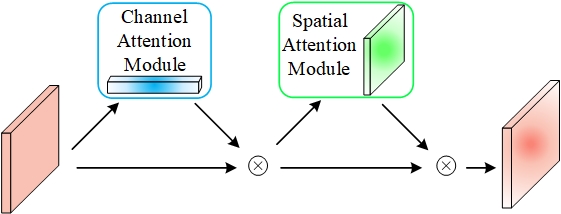}
\caption{Illustration of cross level attention module.}
\label{fig:CLA}
\end{figure*}

\begin{figure*}[!t]
\centering
\subfigure[Channel Attention]{
\includegraphics[scale=0.85]{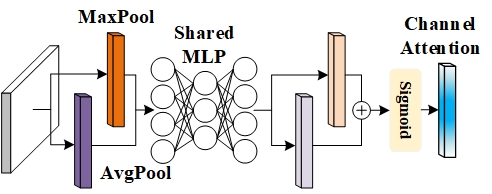}
}
\subfigure[Spatial Attention]{
\includegraphics[scale=0.75]{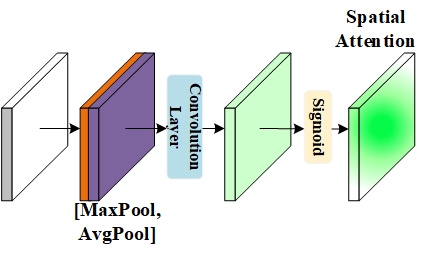}
}
\caption{Two attention sub-module of cross level attention module. CLA is a series connection of channel attention and spatial attention. Both of them uses max-pooling and average-pooling.}
\label{fig:CLA-att}
\end{figure*}

\subsection{Cross Level Attention Module}
In the text detection task, it is important to concatenate the detected character segments into text instances. Feature extractor can easily detect low-level feature such as stroke edge thus localize the position of characters, but the lack of contextual information will make it hard to decide which characters form text instance together. As a result, segmentation based methods often face the problem of over segmentation, which breaks a long text instance into several parts.

To attack the problem mentioned above, we introduce cross level attention module. Inspired by \cite{CBAM}, we adopt channel attention and spatial attention in a sequential manner at the end of the fused feature map. The structure of cross level attention module is shown in Figure.\ref{fig:CLA-att}. In channel attention module, we use max-pooling and average-pooling to squeeze spatial dimension and get two different spatial context descriptors $A_{c}^{max} \in \mathbb{R}^{C\times1\times1}$ and $A_{c}^{avg} \in \mathbb{R}^{C\times1\times1}$. Then the descriptors are forwarded to a multi-layer perceptron to get two feature vectors. Finally we merge the feature vectors with element-wise sum to get channel attention vector $A_{c} \in \mathbb{R}^{C\times1\times1}$. The channel attention is formulated as follow:
\begin{equation}
    A_{c}(F) = \sigma (MLP(AvgPool(F)) + MLP(MaxPool(F)))
\end{equation}
where $\sigma$ denotes the sigmoid function and $MLP$ denotes multi-layer perceptron.

As for spatial attention module, we utilize average-pooling and max-pooling along the channel axis and concatenate them to generate feature descriptor $A_{s}^{avg} \in \mathbb{R}^{1\times H\times W}$ and $A_{s}^{max} \in \mathbb{R}^{1\times H\times W}$. Finally we use convolution layer and sigmoid layer to generate spatial attention map $A_{s} \in \mathbb{R}^{H\times W}$. The spatial attention is computed as:
\begin{equation}
    A_{s}(F) = \sigma (conv([MaxPool(F), AvgPool(F)]))
\end{equation}
where $\sigma$ denotes the sigmoid function, $[\cdot]$ denotes concatenate operation and $conv$ denotes a convolution layer with the kernel size of $7\times7$.

To get features of different scales, we use feature pyramid network to fuse multi-level features in a top-down manner. We argue that, in text detection task, low-level feature needs the guide of contextual information to form text instances, so the idea of multi-level feature fusion operation is to add contextual information to low-level feature. With the help of attention mechanism, model can combine multi-level features to form more accurate text representations.

\subsection{Feature Decomposition-Reconstruction Module}
\begin{figure*}[!t]
\centering
\subfigure[Structure of FDR]{
\label{fig:subfig:a}
\includegraphics[scale=0.67]{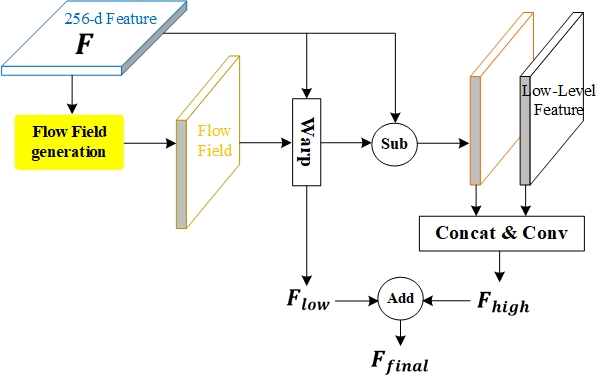}
}
\subfigure[Flow field generation]{
\label{fig:subfig:b}
\includegraphics[scale=0.85]{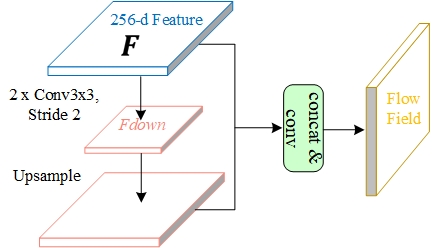}
}
\caption{Illustration of feature decomposition-reconstruction module (a) and differentiable binarization (b). FDR module takes 256-d fused feature as input and enhances text boundary through decomposition and reconstruction. The output of FDR is then fed into DB branch to generate final detection result.}
\label{fig:FDR-DB}
\end{figure*}

Another challenge for segmentation-based methods is that it is difficult for text detectors to distinguish adjacent text instances. To solve this problem, we design a decomposition-reconstruction module which split text features into high frequency term and low frequency term. The structure of FDR is illustrated in Figure.~\ref{fig:FDR-DB}. We consider the low frequency term to contains information inside the text region, while high frequency term contains information of text border region. According to our observation, pixels in text boundary region can help distinguish different text instance, while pixels inside the text region are responsible for text/non-text classification. Our idea is to improve the sensitivity to the text border region by adding low level feature to the decomposed high frequency term. 

In the decomposition-reconstruction operation, we first extract high frequency term from 256-d feature, and then fuse low level feature to it. After that we aggregate fused high frequency term and low frequency term to reconstruct origin feature. To obtain low frequency feature, we first need to learn a flow field $\phi \in \mathbb{R}^{H\times W\times2}$ and then use it to warp the original feature map. Feature maps with low-resolution often contain more low-frequency terms, thus we utilize two $3\times3$ convolutions with stride 2 to downsample $F$ into $F_{down}$.Then we upsample $F_{down}$ to the same resolution as $F$ using bilinear interpolation and concatenate them together and apply a $3\times3$ convolution layer to generate flow field $\phi$. 

The next step is to use $\phi$ to warp $F$. The feature warping procedure can be described as follows. For each position $p$ on feature map $F$, we use flow field $\phi$ to map $p$ to a new point $p'$ via $p' = p + \phi(p)$, then we use bilinear sampling mechanism to approximate each point $p_{x}$ in $F_{low}$. The sampling mechanism interpolates the values of the four nearest neighbor pixel of p, as shown in Equation~\ref{eq:Flow}.
\begin{equation}
    F_{low}(p_{x}) = \sum_{p\in\mathcal{N}(p_{l})}w_{p}F(p)
    \label{eq:Flow}
\end{equation}
where $F$ is the origin feature map and $F_{low}$ is the feature of low-freaquency term, $\mathcal{N}$ is the neighbor pixels of $p_{l}$, $w_{p}$ is bilinear weights on warped spatial grid.

To get high frequency feature, we subtract the low frequency feature map $F_{high}$ from the origin feature $F$. Then we add low-level feature $F_{s}$ with rich detail information to the high frequency feature as a supplement.
Finally, we concatenate them together and use a convolution layer for fusion. The feature fusion procedure can be described as Equation~\ref{eq:Fhigh} as follow. 
\begin{equation}
    F_{high} = conv((F - F_{low}) || F_{s})
    \label{eq:Fhigh}
\end{equation}
where $conv$ is a convolution layer and $||$ is the concatenate operation.

\subsection{Optimization}
We apply supervision on all outputs of DB branch, i.e. probability map, threshold map and binary map. The loss function can be defined as a weighted sum of loss for three maps: 
\begin{equation}
    L = L_{b} + \alpha \times L_{p} + \beta \times L_{t}
\end{equation}
where $L_{p}, L_{b}$ and $L_{t}$ are loss for probability map, binary map and threshold map respectively. We set $\alpha$ to 5 and $\beta$ to 10.

We apply a binary cross-entropy(BCE) loss for $L_{p}$, while for $L_{b}$ we adopt Dice loss~\cite{diceloss}. $L_{t}$ is computed as the sum of $L1$ distances between the prediction and label. To prevent negative samples from dominating the direction of gradient descent, we use hard negative mining in both BCE loss and Dice loss.
\begin{equation}
    L_{p} = \sum_{i \in S_{l}}(y_{i}logx_{i} + (1-y_{i})log(1-x_{i}))
\end{equation}
\begin{equation}
    L_{b} = 1 - 2\frac{\sum_{i \in S_{l}}y_{i}\times x_{i}}{\sum_{i \in S_{l}}|y_{i}| + \sum_{i \in S_{l}}|x_{i}| + \epsilon}
\end{equation}
\begin{equation}
    L_{t} = \sum_{i \in R_{d}}|y_{i}^{*}-x_{i}^{*}|
\end{equation}
where $S_{l}$ is the sampled set with the ratio of positive samples and negative samples is 1 : 3; $R_{d}$ is a set of pixels inside the dilated region $G_{d}$; $y^{*}$ is the label of threshold map.

\section{Experiments}
\label{sec:Exp}
To compare the effectiveness of our model with existing methods, we conduct comprehensive experiments on two benchmark text detection datasets. The experiments include quantitative and qualitative results. We first introduce all datasets used in our experiments, and most of these are commonly used dataset for scene text detection task. Then we give the implementation details of experiment setting and make comparisons with the sate-of-the-art methods on two benchmark datasets. Finally we conduct ablation studies to validate the effectiveness of proposed methods. 

\subsection{Datasets}

{\bf SynthText}~\cite{SynthText} is a large scale synthetic dataset containing 800k images. These images are generated from 8k background images. We use this dataset only for pre-training.

{\bf MSRA-TD500}~\cite{td500} is a multi-language dataset including Chinese and English. It has 300 images for training and 200 images for testing. Text in this dataset are multi-oriented and often has large aspect ratio. Considering the training set is small, we follow the previous works~\cite{EAST,cornerloc,Textsnake} to use extra 400 images from HUST-TR400~\cite{tr400} as training data following previous works.

{\bf Total-Text}~\cite{Totaltext} is a arbitrary shaped dataset including horizontal, multi-oriented and curve text instances. This dataset consists of 1,255 training images and 300 testing images. The annotations are labelled in word-level.

\subsection{Implementation Details}
We use ResNet-50~\cite{resnet50} with DCN~\cite{DCN} as backbone, specifically, we apply modulated deformable convolution in all the $3\times3$ convolution layers in stages conv3, conv4, and conv5 in the ResNet-50. Next to the backbone we adopt FPN~\cite{FPN} with cross level attention module introduced in Chapter.~\ref{sec:Method}. The fused feature is then fed to FDR module to apply decomposition and reconstruction operation. The dimension of low-level feature $F_{s}$ used to supplement high frequency term is set to 48.

%The feature pyramid outputs four groups of feature maps with the channel number all set to 256 and the resolution is 1/4, 1/8, 1/16, 1/32 of the original input image. To apply cross level attention, we first upsample feature maps of size 1/8, 1/16 and 1/32 to 1/4 respectively using nearest neighbor interpolation; and then we adopt channel-wise and saptial-wise attention mechanism on the feature map of size 1/4; finally we reduce the dimension of four groups of feature maps to 64 and concatenate them together to get fused feature.

We first pre-train our models on SynthText dataset for 100k iterations. Then we finetune the models on corresponding datasets for 1200 epochs. Similar to ~\cite{polyLR}, we use the ``poly'' learning rate decay strategy to decrease initial learning rate by multiplying $(1 - \frac{iter}{max\_iter})^{power}$. The initial learning rate is set to 0.007 and $power$ is set to 0.9. Stochastic gradient descent(SGD) is adopted as optimizer with the weight decay of 0.0001 and a Nesterov momentum~\cite{momentum} of 0.9.

In the training phase, images are resized to $640\times640$. Data augmentation strategies like random horizontal flip, random rotation and random crop are applied. We ignore the blurred text instances marked as "DO NOT CARE" in all datasets. The OHEM~\cite{ohem} is adopted with the ratio between negative and positive samples being 3 : 1. During testing, we resize images as follows: For MSRA-TD500, we resize the short edge of input image to 736 while keeping its original aspect ration. For Total-Text, the operation is similar to MSRA-TD500 but the short edge is set to 1152 and 800 respectively.

\subsection{Comparisons with state-of-the-art methods}

\begin{figure*}[!ht]
\centering
\includegraphics[scale=0.4]{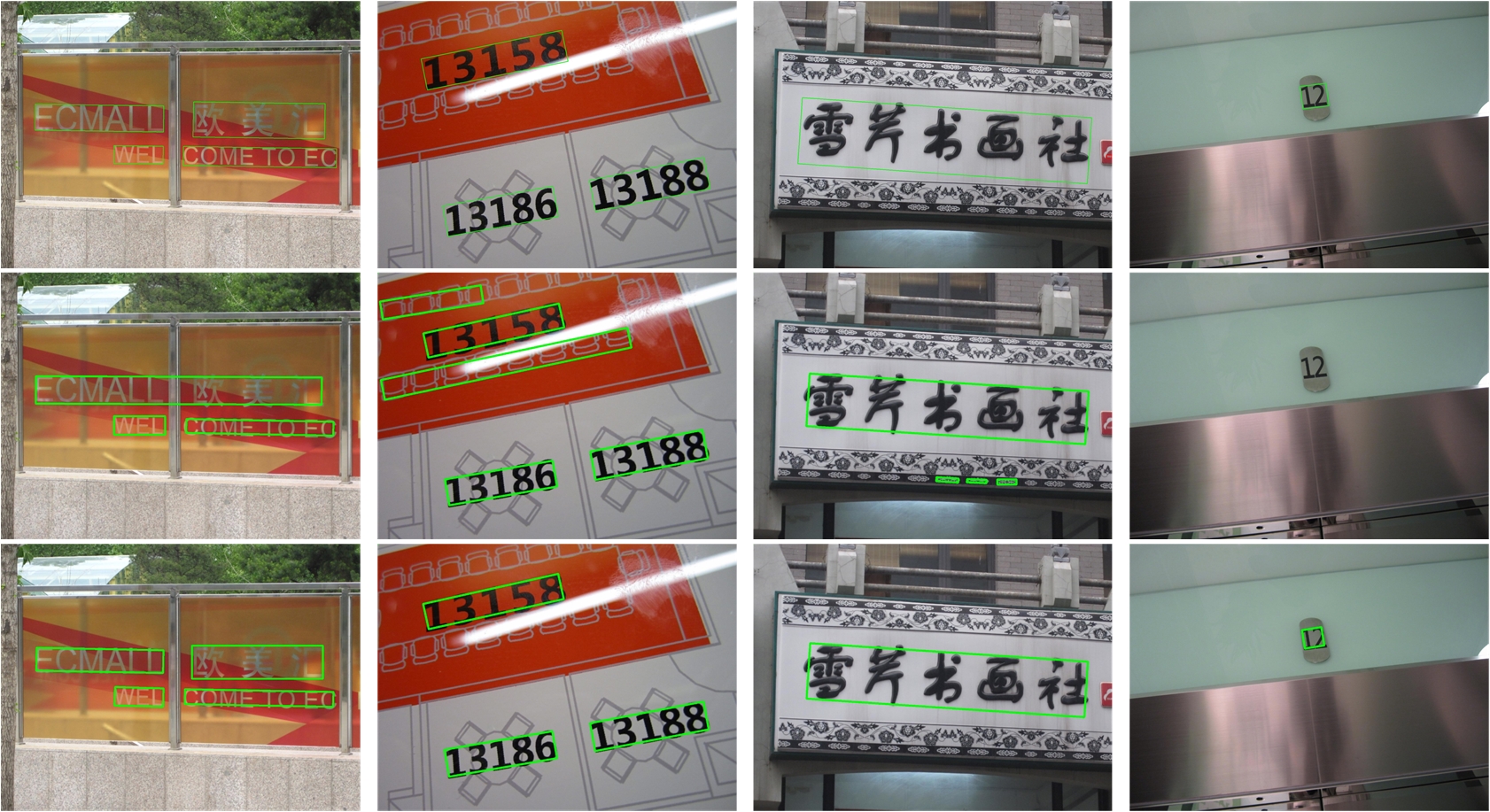}
\caption{Visualization of detection results on MSRA-TD500 dataset. First row is ground-truth. Second row is output of DB and third row is output of FDRNet. Text region marked with red is ``DO NOT CARE".}
\label{fig:vis-td500}
\end{figure*}

\begin{figure*}[!ht]
\centering
\includegraphics[scale=0.4]{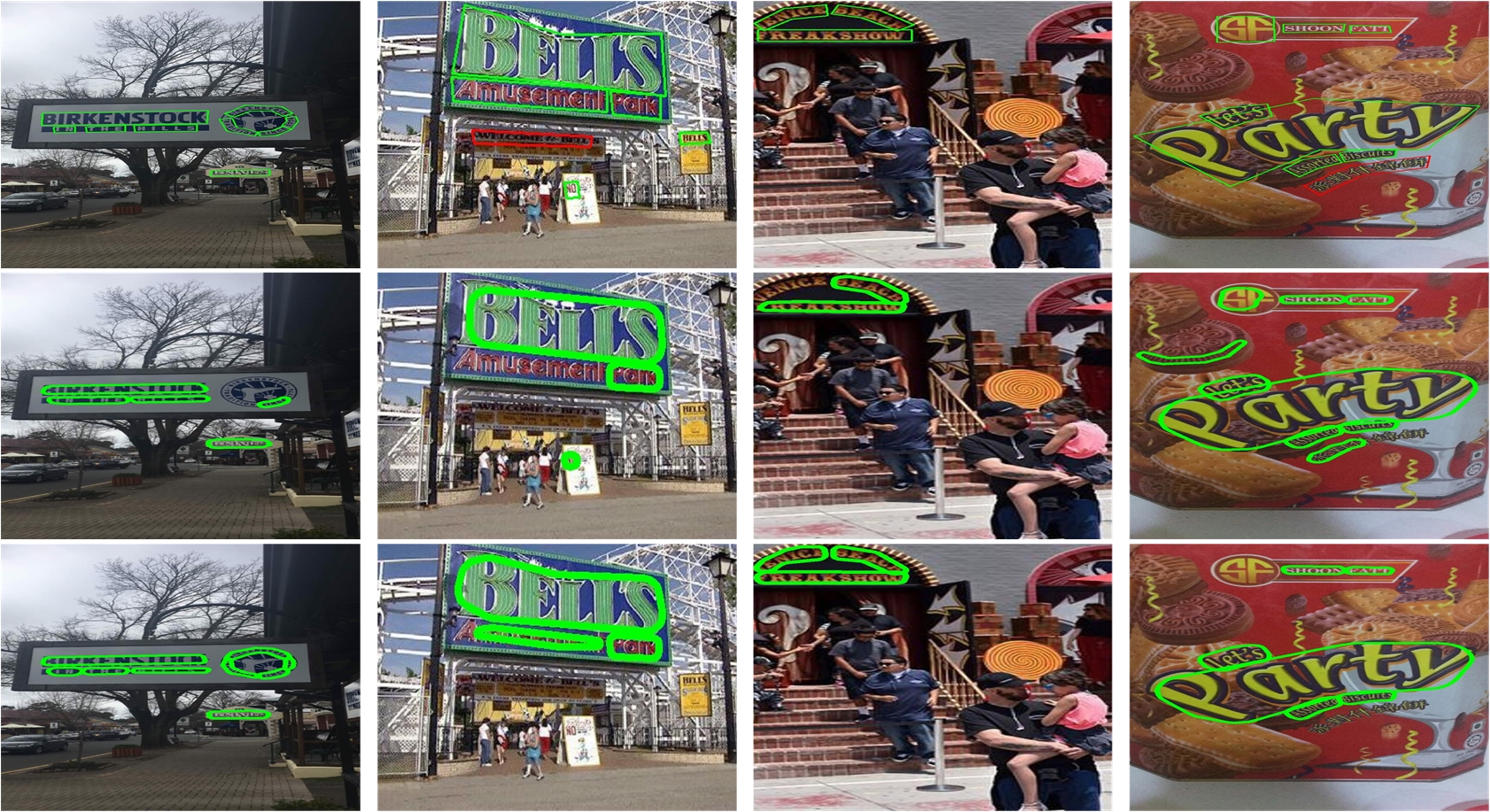}
\caption{Visualization of detection results on Total-text dataset. First row is ground-truth. Second row is output of DB and third row is output of FDRNet. Text region marked with red is ``DO NOT CARE".}
\label{fig:vis-total-text}
\end{figure*}

We compare FDRNet with previous state-of-the-art text detectors on MSRA-TD500 and Total-Text datasets. Unless mentioned otherwise, we did not use extra data or tricks like multi-scale testing for fair comparison. Quantitative results are given in Tables~\ref{tab:compare-td500} and ~\ref{tab:compare-total-text}.

Figure~\ref{fig:vis-td500} and Figure~\ref{fig:vis-total-text} are qualitative comparisons between FDRNet and DB on MSRA-TD500 and Total-text dataset respectively. Visualization results show that FDRNet's detection result are very close to the ground-truth. However, problems of text adhesion, over segmentation, missed detection and false alarm often occur in DB's results.

\begin{table}[!ht]
\caption{Comparisons with previous text detection algorithms on MSRA-TD500 dataset.}
\begin{tabular}{p{1.8cm}p{1.8cm}p{1.8cm}p{1.8cm}cccc}
\hline
Method    & Precision & Recall & F-score \\ \hline
EAST~\cite{EAST}      & 87.3          & 67.4       & 76.1    \\
SegLink~\cite{SegLink}   & 86.0          & 70.0       & 77.0    \\
PixelLink~\cite{PixelLink} & 83.0          & 73.2       & 77.8    \\
TextSnake~\cite{Textsnake} & 83.2          & 73.9       & 78.3    \\
TextField~\cite{Textfield} & 87.4          & 75.9       & 81.3    \\
ATRR~\cite{ATRR}      & 85.2          & 82.1       & 83.6    \\
MSR~\cite{MSR}       & 87.4          & 76.7       & 81.7    \\
SAE~\cite{SAE}       & 84.2          & 81.7       & 82.9    \\
DRRG~\cite{DRRG}      & 88.1          & 82.3       & 85.1    \\
PAN~\cite{PAN}       & 85.7          & \textbf{83.2}       & 84.5    \\
DB~\cite{DB}        & 91.5          & 79.2       & 84.9    \\
CRAFT~\cite{CRAFT}     & 88.2          & 78.2       & 82.9    \\
MOST~\cite{MOST}      & 90.4          & 82.7       & 86.4    \\ \hline
ours      & \textbf{92.5}          & 82.3       & \textbf{87.1}    \\ \hline
\end{tabular}
\label{tab:compare-td500}
\end{table}

On the MSRA-TD500 dataset, FDRNet achieved the highest precision and F-score. It surpassed DB by 1.0\%, 2.9\% and 2.2\% on precision, recall and F-score respectively. Such good performance mainly depends on high precision, which means that the incidence of false alarm and text adhesion is low. As an arbitrary shaped text detector, our model performances even better than multi-oriental text detector MOST, which indicates the overwhelming advantage in such task.

\begin{table}[!ht]
\caption{Comparisons with previous text detection algorithms on Total-Text dataset.}
\begin{tabular}{p{1.8cm}p{1.8cm}p{1.8cm}p{1.8cm}cccc}
\hline
Method     & Precision & Recall & F-score \\ \hline
TextSnake~\cite{Textsnake}  & 82.7      & 74.5   & 78.4    \\
TextField~\cite{Textfield}  & 81.2      & 79.9   & 80.6    \\
ATRR~\cite{ATRR}       & 80.9      & 76.2   & 78.5    \\
PSENet~\cite{PSENet}     & 84.0      & 78.0   & 80.9    \\
ContourNet~\cite{ContourNet} & 86.9      & \textbf{83.9}   & 85.4    \\
LOMO~\cite{LOMO}       & 88.6      & 75.7   & 81.6    \\
PAN~\cite{PAN}        & \textbf{89.3}      & 81.0   & 85.0    \\
DB~\cite{DB}         & 87.1      & 82.5   & 84.7    \\
CRAFT~\cite{CRAFT}      & 87.6      & 79.9   & 83.6    \\
FCENet~\cite{FCENet}     & \textbf{89.3}      & 82.5   & 85.8    \\ \hline
ours       & 88.4      & 83.7   & \textbf{86.0}    \\ \hline
\end{tabular}
\label{tab:compare-total-text}
\end{table}

On the Total-Text dataset, FDRNet achieves the highest F-score by 86.0\%. Though our proposed method did not get best results in precision or recall, but both of them are at first class level and jointly make F-score surpass FCENet by 0.2\%. Compared with ContourNet, the advantage of our method is high precision. Experiments on this dataset demonstrate the robustness of FDRNet in detection of arbitrary shaped text.

\subsection{Ablation Study}
\begin{figure}[!ht]
\centering
\includegraphics[scale=0.37]{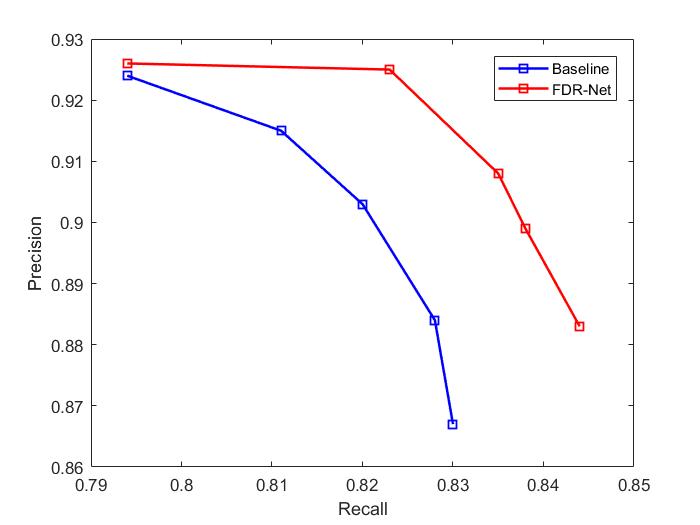}
\caption{Comparison of PR curve on MSRA-TD500 dataset. The blue line represents baseline model and the red line represents FDRNet.}
\label{fig:PRCurve}
\end{figure}

To compare the performance of FDRNet and baseline model DB, we draw PR curve by setting different IoU threshold on MSRA-TD500 dataset as shown in Figure~\ref{fig:PRCurve}. We can find that FDRNet has larger area-under-curve than its counterpart DB.

\textbf{Combination of FDR and CLA} To validate the effectiveness of CLA module and FDR module, we conduct ablation study on MSRA-TD500 dataset and Total-Text dataset as shown in Table \ref{tab:ablation-td500} and Table \ref{tab:ablation-total-text}. The result of baseline model is generated by re-implement DB. On both datasets, model with FDR and CLA improves both precision and recall thus achieved the best results and surpassed the baseline model by a large margin. Moreover, independently using CLA or FDR module can also bring improvement as well.

\begin{table}[]
\caption{Evaluation of FDRNet with or without CLA and FDR module on MSRA-TD500 dataset.}
\begin{tabular}{p{1.8cm}p{1.8cm}p{1.8cm}p{1.8cm}cccc}
\hline
Method    & Precision & Recall & F-score \\ \hline
Baseline  & 89.8      & 80.2   & 84.8    \\
+FDR      & 90.9      & 82.1   & 86.3    \\
+CLA      & 92.1      & 80.6   & 86.0    \\ 
+FDR+CLA  & 92.5      & 82.3   & 87.1    \\ \hline
\end{tabular}
\label{tab:ablation-td500}
\end{table}

\begin{table}[]
\caption{Evaluation of FDRNet with or without CLA and FDR module on Total-Text dataset.}
\begin{tabular}{p{1.8cm}p{1.8cm}p{1.8cm}p{1.8cm}cccc}
\hline
Method    & Precision & Recall & F-score \\ \hline
Baseline  & 87.4      & 80.8   & 84.0    \\
+FDR      & 87.4      & 81.3   & 84.2    \\
+CLA      & 87.6      & 81.8   & 84.6    \\ 
+FDR+CLA  & 88.4      & 83.7   & 86.0    \\ \hline
\end{tabular}
\label{tab:ablation-total-text}
\end{table}

\textbf{Number and position of CLA module } To study how the number and position of CLA module affect model performance, we trained FDRNet containing 1, 2 and 4 CLA modules respectively. From Table~\ref{tab:ablation-CLA}, we can see that model with only one CLA reaches the best result. That is because the high resolution feature map out2 has relatively smaller receptive field and rich low-level features, which is helpful to locate text region. Adding CLA modules on low resolution feature map out3, out4 and out5 will increase the cost of learning, but the information can be used by attention mechanism is limited, thus resulting in performance degradation. 

\begin{table}[!ht]
\caption{Evaluation of FDRNet with different number and position of CLA module on MSRA-TD500 dataset. CLA-1 represents model with one CLA module connected after layer out2. CLA-2 for two CLA at out2 and out3 while CLA-4 for four CLA at out2, out3, out4 and out5.}
\begin{tabular}{p{1.8cm}p{1.8cm}p{1.8cm}p{1.8cm}cccc}
\hline
Method    & Precision & Recall & F-score \\ \hline
CLA-1  & 92.5      & 82.3   & 87.1    \\
CLA-2      & 89.5      & 81.8   & 85.5    \\
CLA-4  & 90.5      & 80.4   & 85.2    \\ \hline
\end{tabular}
\label{tab:ablation-CLA}
\end{table}

\textbf{Low-level feature of FDR module} To study how different selection of low-level feature in FDR module affect model performance, we trained FDRNet choosing conv2 and conv3 of backbone as low-level feature in FDR module respectively. From Table~\ref{tab:ablation-FDR}, we can see that choosing conv2 as low-level feature performs better than conv3. We consider that the use of conv3 needs additional upsample operation and it will introduce low-frequency noise, while conv2 contains richer detail information so is more suitable for feature fusion.

\begin{table}[!ht]
\caption{Evaluation of FDRNet with different low-level feature selection strategy in FDR module on MSRA-TD500 dataset.}
\begin{tabular}{p{1.8cm}p{1.8cm}p{1.8cm}p{1.8cm}cccc}
\hline
Method    & Precision & Recall & F-score \\ \hline
FDR-conv2  & 92.5      & 82.3   & 87.1    \\
FDR-conv3  & 92.9      & 81.4   & 86.8    \\ \hline
\end{tabular}
\label{tab:ablation-FDR}
\end{table}

% In comparison to the baseline model, the introduction of either the Cross Level Attention module or the Feature Decomposition-Reconstruction module increases the accuracy of the detection result. Combining both modules can further improve the performance of text detector. The experimental results show that our proposed method which adopt both CLA and FDR achieves the best, outperforming the baseline by 2.3\% and 1.3\% on MSRA-TD500 and Total-Text dataset respectively. Specifically, both CLA and FDR improves the precision and recall of baseline model. By adopting FDR module, we achieve recall of 82.1\% and f-measure of 86.3\% on MSRA-TD500, outperforming the baseline by 1.9\% and 1.5\% respectively. On the other hand, the CLA module can also brings performance gain by 2.3\% on precision and 1.2\% on f-measure, since it can filters text features from cluttered background features. Quantitative and qualitative result on MSRA-TD500 is shown in Figure~\ref{fig:quant-quali}.

\subsection{Failure cases analysis}
In order to study why our proposed method fails on MSRA-TD500 and Total-Text dataset, we show FDRNet's intermediate results to help diagnosing failure cases. Though the final result is generated only by probability map, threshold map also participates in optimization process during training phase, so it can also reflect the motivation of network to make decision. 

\begin{figure*}[!ht]
\centering
\includegraphics[scale=0.5]{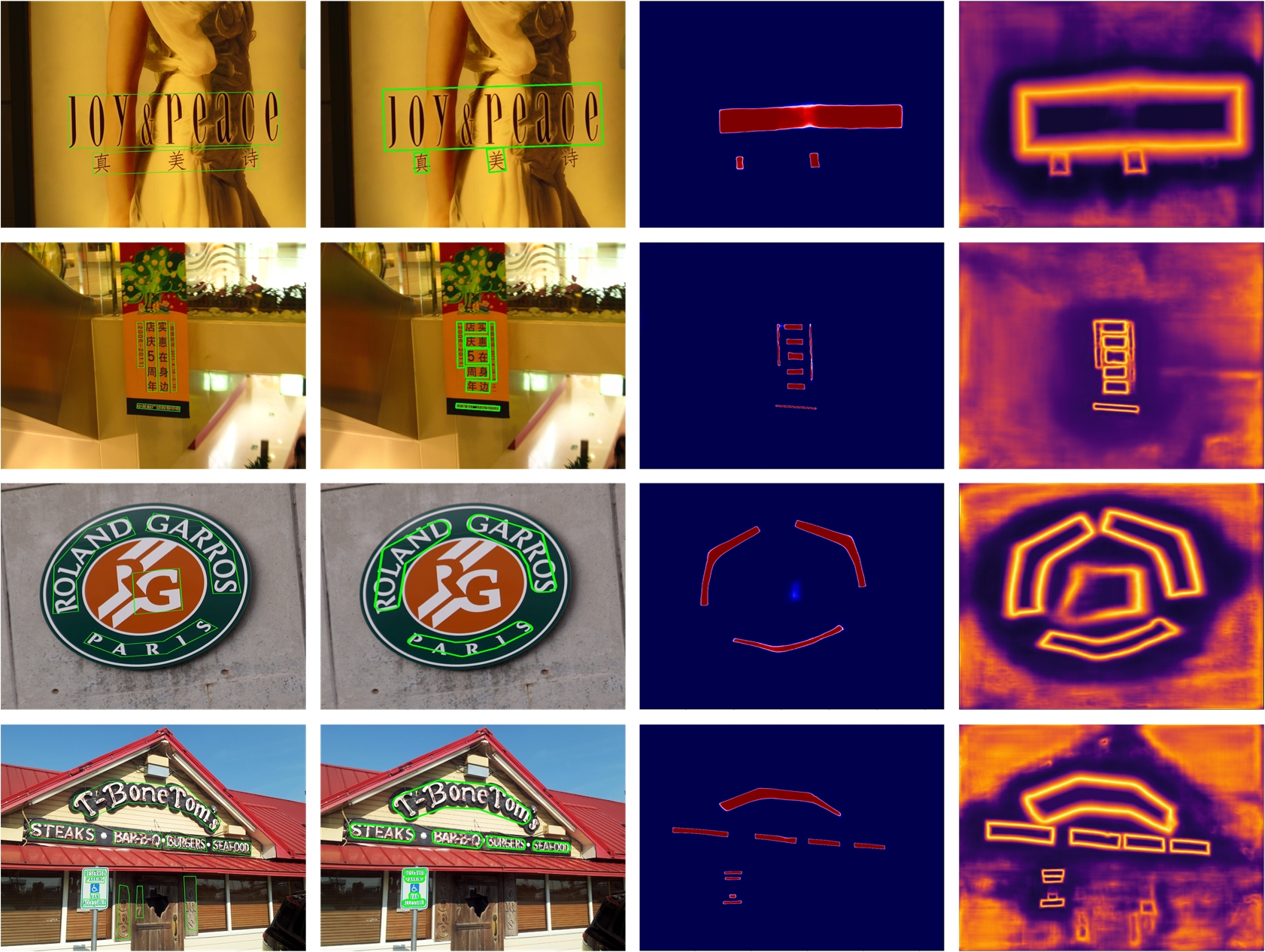}
\caption{Failure results of FDRNet on MSRA-TD500 and Total-Text dataset. From left to right, ground-truth, detection result, probability map and threshold map. }
\label{fig:vis-bad-results}
\end{figure*}

As shown in Figure~\ref{fig:vis-bad-results}, text region perceived by network is slightly different on probability map and threshold map. The first line shows problem of wrong detection. It is main caused by three chinese characters with extremly large text spacing. The second line shows our model wrongly detect vertical text into horizontal text. Such phenomenon is quiet common when most text instances in dataset appear as horizontal text. We think an end-to-end model which integrate text detection and recognition model can relieve this problem through semantic information obtained by recognition model. The third line represents the failure of detecting individual character, it is caused by lack of information inside the text as threshold map predicts better than probability map. The fourth line shows when the color and texture of text are similar to background, both probability map and threshold map fails to find text region. 

\subsection{Grad-CAM Visualization}
To analyze the impact of our proposed CLA and FDR module on baseline model from feature level, we visualize the heat map inspired by gradient-weighted class activation map(Grad-CAM). Grad-CAM originated from image classification, in which task the last layer of neural network outputs an n-dimensional probability vector to represent the categories of objects in the test image. Grad-CAM uses classification results to do back propagation and obtain gradient of the last convolution layer as the weight of the feature map. By calculating the weight-sum of feature map, we can get heat map which represent the importance of pixels at different spatial positions on the feature map.as shown in Equation~\ref{eq:grad-cam-weight}.

\begin{equation}
    \alpha_{k}^{c} = \frac{1}{Z}\sum_{i}\sum_{j}\frac{\partial y^{c}}{\partial A_{ij}^{k}}
    \label{eq:grad-cam-weight}
\end{equation}

where $y^{c}$ is the confidence score of class c, $A_{ij}^{k}$ is the pixel of index $(i,j)$ on feature map $A^{k}$, $Z$ is the number of pixels. After getting the weight $\alpha_{k}^{c}$, we use it to calculate the heat map as shown in Equation~\ref{eq:grad-cam-heatmap}.

\begin{equation}
    L_{Grad-CAM}^{c} = ReLU(\sum_{k}\alpha_{k}^{c}A^{k})
    \label{eq:grad-cam-heatmap}
\end{equation}

\begin{figure*}[!t]
\centering
\includegraphics[scale=0.5]{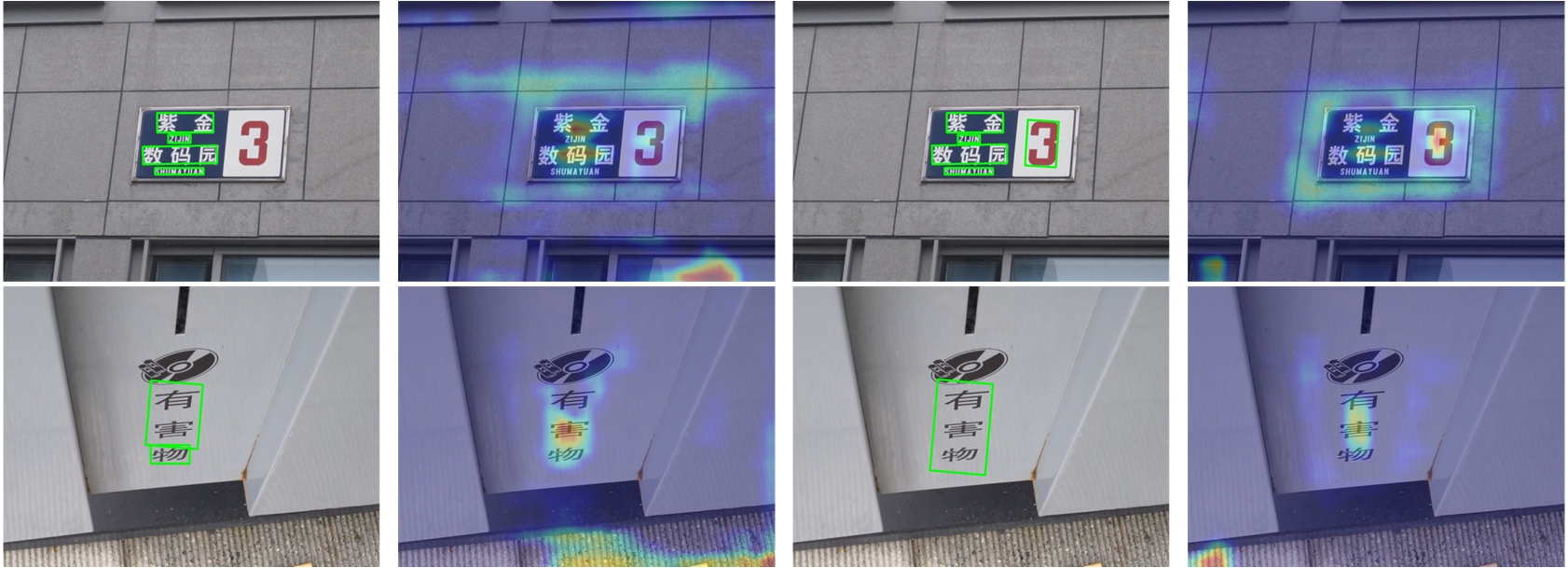}
\caption{Visual comparison of Grad-CAM heat map on MSRA-TD500 test set. From left to right, detection result of baseline, heat map of baseline, detection result of FDRNet, heat map of FDRNet.}
\label{fig:grad-cam-td500}
\end{figure*}

\begin{figure*}[!t]
\centering
\includegraphics[scale=0.5]{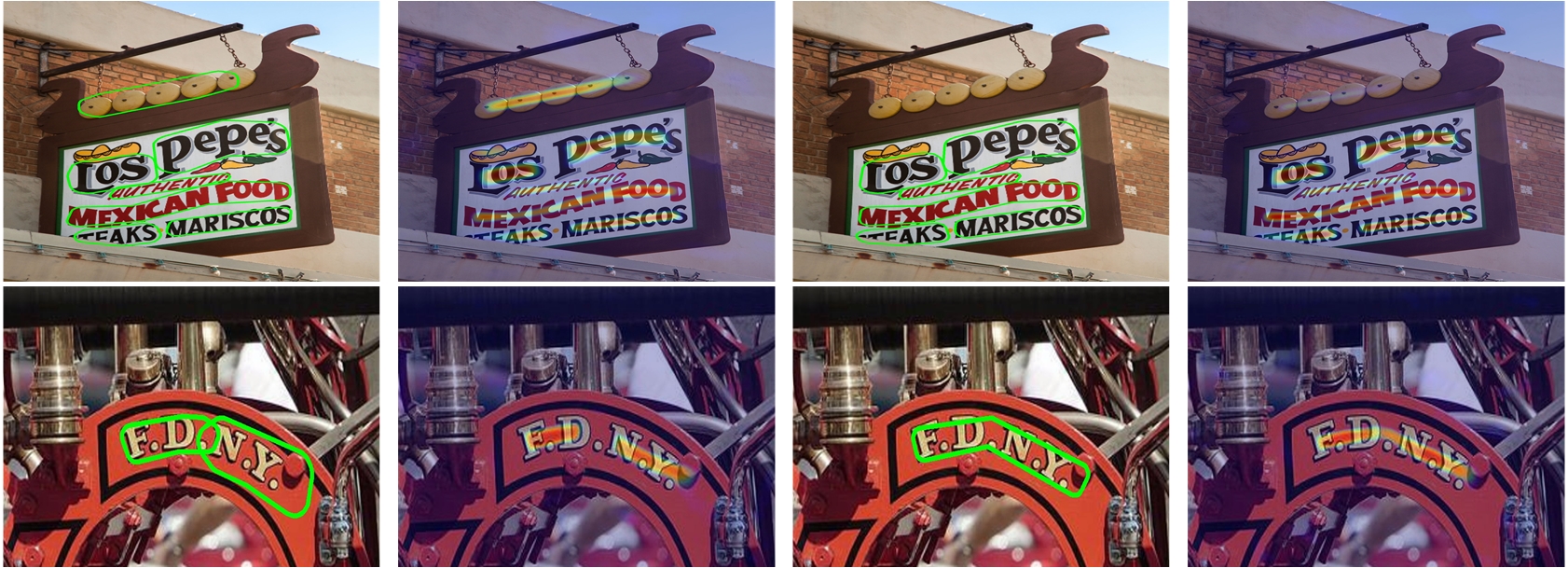}
\caption{Visual comparison of Grad-CAM heat map on Total-Text test set. From left to right, detection result of baseline, heat map of baseline, detection result of FDRNet, heat map of FDRNet}
\label{fig:grad-cam-total-text}
\end{figure*}

The ReLU function is used to select the pixels in feature map which have positive impact on classification result. Such operation can make heat map only focus on the feature which contributes to decision making. Since the output of semantic segmentation model is a probability map of which every pixel represents a class and relevant confidence score, we adjust grad-cam algorithm to generate heat map for segmentation result. Specifically, the probability map only have 1 channel, representing the confidence a pixel belongs to text area, so we sum up all pixels of the image to get the value used for back propagation. For MSRA-TD500 dataset, we choose the last convolutional layer of ResNet-50 stage 4 to generate heat map. As for Total-Text dataset, we select the last convolutional layer of the entire model due to the smaller resolution. The visualization result is shown in Figure.~\ref{fig:grad-cam-td500} and Figure.~\ref{fig:grad-cam-total-text}.

The first line of Figure.~\ref{fig:grad-cam-td500} shows that the baseline model does not make full use of the center of text region for prediction, while in our model this region is significantly enhanced. The second line of Figure.~\ref{fig:grad-cam-total-text} shows that our model can performance better on filtering background area, thus reduce false alarm.

Visualization result also shows the ability of our proposed method to deal with over segmentation. In the second line of Figure.~\ref{fig:grad-cam-td500} and first line of Figure.~\ref{fig:grad-cam-total-text}, the activation area of pixel inside text instance is discontinuous, which lead to over segmentation. Our model can better integrate these detected text fragments.

\section{Conclusion}
In this paper, we propose an Attention Based Feature Decomposition-Reconstruction Network for scene text detection. The cross level attention module in our network is designed to enrich contextual information and reduce over segmentation. The feature decomposition-reconstruction module in our network aims at decomposing features into high frequency term and low frequency term. By merging low level feature into the high frequency term, the output feature map contains more information of text boundary, which is helpful to alleviate the problem of text adhesion. Experiments on various benchmarks demonstrate that our method can achieve a performance comparable to state-of-the-art methods for both multi-oriented text and arbitrary shaped text.

\section{Acknowledgments}
This work was supported by the National Natural Science Foundation of China~(No.62072021).
% Use the style of numbering in square brackets.
% If nothing is used, default style will be taken.
%\begin{enumerate}[a)]
%\item 
%\item 
%\item 
%\end{enumerate}  

%\begin{itemize}
%\item 
%\item 
%\item 
%\end{itemize}  

%\begin{description}
%\item[]
%\item[] 
%\item[] 
%\end{description}  

% Figure
% \begin{figure}[<options>]
% 	\centering
% 		\includegraphics[<options>]{}
% 	  \caption{}\label{fig1}
% \end{figure}

% \begin{table}[<options>]
% \begin{tabular*}{\tblwidth}{@{}LL@{}}
% \toprule
% \midrule
%  & \\
%  & \\
%  & \\
%  & \\
% \bottomrule
% \end{tabular*}
% \end{table}

% Uncomment and use as the case may be
%\begin{theorem} 
%\end{theorem}

% Uncomment and use as the case may be
%\begin{lemma} 
%\end{lemma}


\printcredits

%\bibliographystyle{cas-model2-names}

\bibliography{cas-refs}

\bio{}
\endbio

\endbio

\end{document}